\documentclass{article}

\PassOptionsToPackage{numbers}{natbib}
\PassOptionsToPackage{sort,compress}{natbib}
\usepackage[colorlinks,bookmarksopen,bookmarksnumbered,citecolor=green, linkcolor=red]{hyperref}
\usepackage[final]{neurips_2024}

\usepackage[utf8]{inputenc} 
\usepackage[T1]{fontenc}    
\usepackage{hyperref}       
\usepackage{url}            
\usepackage{booktabs}       
\usepackage{amsfonts}       
\usepackage{nicefrac}       
\usepackage{microtype}      
\usepackage{xcolor}         
\usepackage{graphicx}
\usepackage{algorithm}
\usepackage{algpseudocode}
\usepackage{amsmath}
\usepackage{makecell}
\usepackage{multirow}
\usepackage{multicol}
\newcommand{\blackhref}[2]{\href{#1}{\textcolor{black}{#2}}}
\makeatletter
\def\@fnsymbol#1{\ensuremath{\ifcase#1\or \textcolor{black}{*}\or \textcolor{black}{**}\or \textcolor{black}{***}\else \@ctrerr \fi}}
\makeatother

\title{SA3DIP: Segment Any 3D Instance with \\ Potential 3D Priors}

\author{
Xi Yang\textsuperscript{1}, Xu Gu\textsuperscript{1}, Xingyilang Yin\textsuperscript{1}\thanks{Corresponding author.} , Xinbo Gao\textsuperscript{2} \\
\textsuperscript{1}Xidian University, \textsuperscript{2}Chongqing University of Posts and Telecommunications \\
\texttt{yangx@xidian.edu.cn}, \texttt{\{ryangu,yxyl\}@stu.xidian.edu.cn}, \texttt{gaoxb@cqupt.edu.cn}\\
}

\begin{document}

\maketitle

\begin{abstract}
	The proliferation of 2D foundation models has sparked research into adapting them for open-world 3D instance segmentation. Recent methods introduce a paradigm that leverages superpoints as geometric primitives and incorporates 2D multi-view masks from Segment Anything model (SAM) as merging guidance, achieving outstanding zero-shot instance segmentation results. However, the limited use of 3D priors restricts the segmentation performance. Previous methods calculate the 3D superpoints solely based on estimated normal from spatial coordinates, resulting in under-segmentation for instances with similar geometry. Besides, the heavy reliance on SAM and hand-crafted algorithms in 2D space suffers from over-segmentation due to SAM's inherent part-level segmentation tendency. To address these issues, we propose SA3DIP, a novel method for \textbf{S}egmenting \textbf{A}ny \textbf{3D} \textbf{I}nstances via exploiting potential 3D \textbf{P}riors. Specifically, on one hand, we generate complementary 3D primitives based on both geometric and textural priors, which reduces the initial errors that accumulate in subsequent procedures. On the other hand, we introduce supplemental constraints from the 3D space by using a 3D detector to guide a further merging process. Furthermore, we notice a considerable portion of low-quality ground truth annotations in ScanNetV2 benchmark, which affect the fair evaluations. Thus, we present ScanNetV2-INS with complete ground truth labels and supplement additional instances for 3D class-agnostic instance segmentation. Experimental evaluations on various 2D-3D datasets demonstrate the effectiveness and robustness of our approach. Our code and proposed ScanNetV2-INS dataset are available \blackhref{https://github.com/RyanG41/SA3DIP}{HERE}.
\end{abstract}

\section{Introduction}

3D instance segmentation is a fundamental task pivotal to 3D understanding across various domains such as autonomous driving, robotics navigation, and virtual reality applications. State-of-the-art methods \cite{shin2024spherical,schult2023mask3d} are predominantly supervised and rely heavily on precise 3D annotations for training, thus constraining their applications in open-world scenes. Compared to scarce 3D labeled data, the acquisition and annotation of 2D images are more convenient. Recently, 2D foundation models \cite{kirillov2023segment,radford2021learning,caron2021emerging,oquab2023dinov2} trained on large-scale annotated 2D data show impressive performance and strong generalization capabilities in zero-shot scenarios. Recent efforts have sought to leverage Segment Anything Model (SAM) by lifting its class-agnostic 2D segmentation results to 3D tasks \cite{guo2023sam-graph,yin2023sai3d,xu2023sampro3d,yang2023sam3d}. Specifically, some methods \cite{yin2023sai3d,guo2023sam-graph} propose a pipeline that decomposes the 3D scene into geometric primitives and leverages 2D multi-view masks from SAM to calculate pairwise similarity scores as merging guidance. Further well-designed algorithms or Graph Neural Networks (GNNs) are included to ensure multi-view consistency.

\begin{figure}
	\centering
	\includegraphics[width=\linewidth]{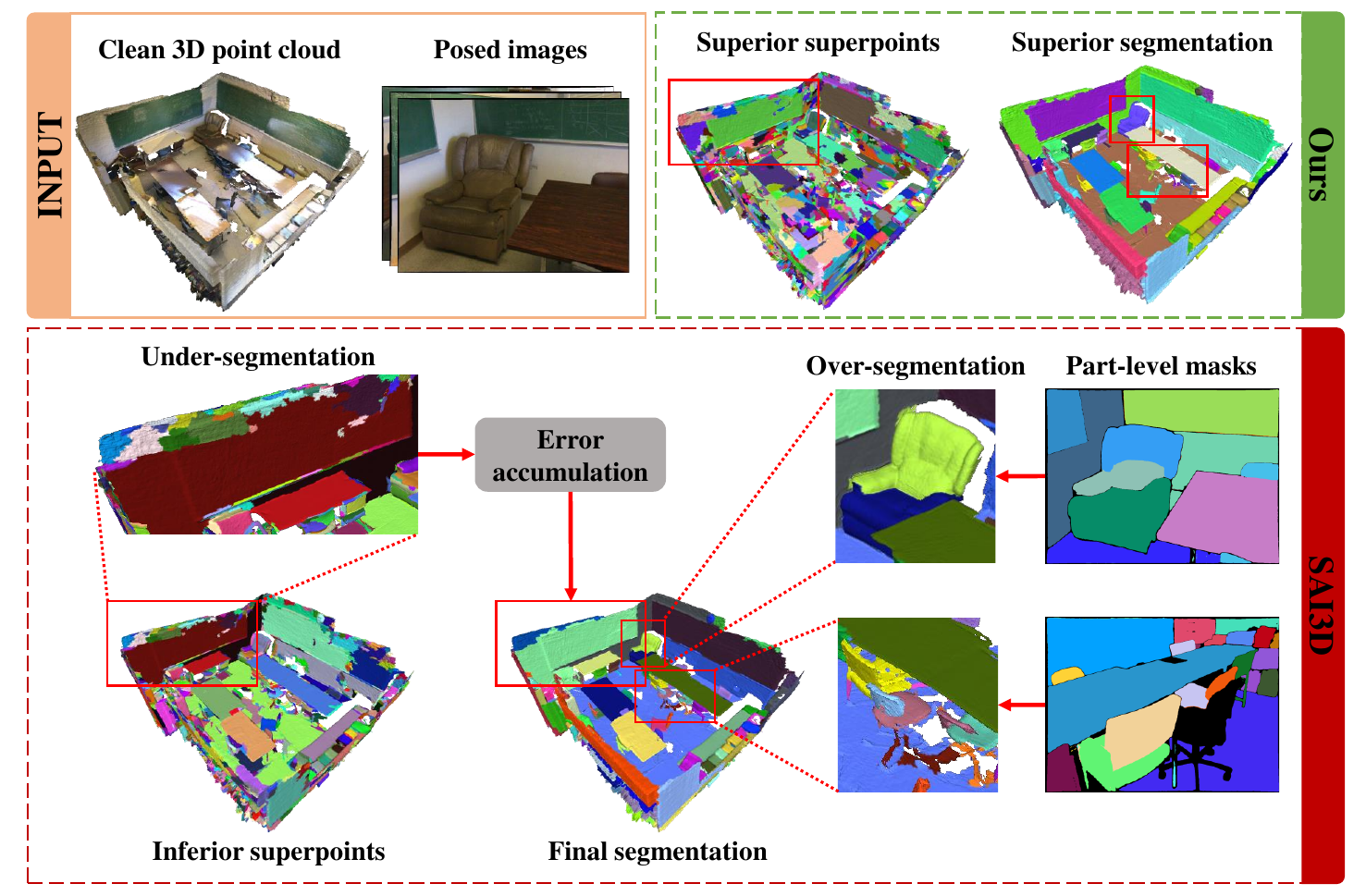}
	\caption{Comparison of our \textbf{SA3DIP} with other methods. Methods like SAI3D (bottom) fail to distinguish instances with similar normals when computing superpoints, which accumulate to the final segmentation. Moreover, the part-level 2D segmentation transfers to 3D space, resulting in over-segmented 3D instances. We present a novel pipeline for segmenting any 3D instances, which overcomes the limitations by exploiting additional 3D priors, specifically by incorporating both geometric and textural prior on superpoints computing, and supplementing 3D space constraint provided 3D prior by utilizing a 3D detector.}
	\label{figure1:intro}
\end{figure}

However, the geometric rudimentary pre-segmentation initialization impedes their ability to group superpoints on points with highly similar normals, such as boards on walls and books on tabletops. As shown in Fig. \ref{figure1:intro} bottom left, the blackboard and the wall are wrongly allocated within the same superpoint using previous methods. Owing to the coarse-to-fine pattern of the pipeline, errors at this stage propagate to subsequent stages, which the sophisticated merging algorithms fail to rectify. Furthermore, current approaches heavily rely on 2D foundation models and design algorithms or GNNs within 2D space, neglecting the inherent 3D priors of the data. Part-level segmentation in the generated 2D masks by SAM transfers to 3D space and leads to over-segmented 3D instances. As illustrated in Fig. \ref{figure1:intro} bottom right, the sofa and chairs are segmented at the part level in 2D space, causing over-segmentation in the final results. These limitations primarily stem from the under-exploitation of 3D priors: (1) Complete point cloud data encompasses not only spatial coordinates but also color channels; (2) Constraints provided by 3D space prior to the merging process cannot be neglected.

In this paper, we present \textbf{SA3DIP} (\textbf{S}egment \textbf{A}ny \textbf{3D} \textbf{I}nstance with potential 3D \textbf{P}riors), a novel method for segmenting high-quality 3D instances. Specifically, we observe that distinct instances with similar normals often exhibit different colors. Therefore, we incorporate both geometric and textural priors to generate finer-grained complementary primitives. As shown in Fig. \ref{figure1:intro} top right, our method identifies the boundary between the blackboard and the wall clearly. In this way, the initial errors are minimized, which reduces error accumulation in the subsequent process. Moreover, we exploit the 3D prior at the merging stage to provide constraints on the over-segmented 3D instances, which is implemented by incorporating a 3D detector. This additional 3D prior enables rectification on the over-segmented 3D instances, while preserving the capability in handling fine-grained objects. Therefore, the sofa and chairs maintain their integrity in 3D space by our approach, which is illustrated in Fig. \ref{figure1:intro} top right. Additionally, we notice that the widely-used benchmark, ScanNet \cite{dai2017scannet}, contains a considerable portion of low-quality ground truth annotations for instance segmentation, which leads to biases in assessing model performance. Thus, we propose ScanNetV2-INS, a point-level enhanced version tailored for 3D class-agnostic instance segmentation. The revised dataset contains fewer incomplete labels and fewer missing instances, which better showcases real-world scenarios.

Our contributions are three-fold: (1) We present \textbf{SA3DIP}, a novel pipeline for segmenting any 3D instances by exploiting potential 3D priors, which includes incorporating both geometric and color priors on computing 3D superpoints, and introducing constraints provided by 3D prior at the merging stage; (2) We propose a point-level enhanced version of ScanNetV2, specifically for 3D class-agnostic instance segmentation by rectifying incomplete annotations and incorporating more instances; (3) Extensive experiments are conducted on ScanNetV2, ScanNetV2-INS and ScanNet++ \cite{yeshwanthliu2023scannetpp} datasets, and the competitive results demonstrate the effectiveness and robustness of our method.

\section{Related Work}

\paragraph{Close-set 3D segmentation.}
3D semantic segmentation aims to classify each point into a specific semantic class \cite{qi2017pointnet,qi2017pointnet++,graham20183d,xu2021paconv,qian2022pointnext,yin2024point,zhao2021point,wu2022point,wu2023point}. 3D instance segmentation, on the other hand, assigns unique masks to each distinct instance within the same semantic category \cite{schult2023mask3d,choy20194d,liang2021instance,sun2023superpoint,takmaz2024openmask3d,thomas2019kpconv}. Prior research, categorized as Grouping-based \cite{chen2021hierarchical,jiang2020pointgroup,liang2021instance}, Kernel-based \cite{wu20223d}, and Transformer-based \cite{lai2023mask,schult2023mask3d,lu2023query} methods, has primarily relied on labeled datasets in a supervised manner. Mask3D \cite{schult2023mask3d} proposes the first Transformer-based model for 3D semantic instance segmentation that uses instance queries and Transformer decoders. Spherical Mask \cite{shin2024spherical} achieves state-of-the-art 3D instance segmentation performance on the ScanNetV2 dataset by leveraging a novel coarse-to-fine approach \cite{hou20193d,kolodiazhnyi2024top} based on spherical representation. However, they all necessitate a significant corpus of annotated 3D data for network training, which is financially burdensome and poses challenges for extending the method to open-world scenarios featuring novel objects from unobserved categories.

\paragraph{Open-set 3D segmentation.}
2D foundation models \cite{kirillov2023segment,radford2021learning,caron2021emerging,oquab2023dinov2} have exhibited remarkable efficacy across various tasks. Training on the SA-1B dataset with 11 million images and 1.1 billion masks, Segment Anything model (SAM) serves as a cornerstone for image segmentation, allowing strong zero-shot transfer ability and diverse prompts such as points, boxes, and texts to generate high-quality segmentation masks. Inspired by the generalization capabilities of foundation models, certain works \cite{yang2023sam3d,guo2023sam-graph,yin2023sai3d,xu2023sampro3d,liu2024segment,ding2023pla,yang2023regionplc,lu2023ovir} explore the feasibility of bridging the gap between 2D and 3D, enabling various open-vocabulary 3D tasks. OpenScene \cite{peng2023openscene} and OpenMask3D \cite{takmaz2024openmask3d} both rely on transferring knowledge from CLIP, where the former infers CLIP features for each 3D point and classifies them embeddings of class labels, and the latter uses additional pre-trained 3D segmentation model to produce class-agnostic 3D proposals and classifies them based on CLIP scores. SAM3D \cite{yang2023sam3d} pioneers the extension of SAM into 3D perception by transferring segmentation information from 2D images to 3D space. SAMPro3D \cite{xu2023sampro3d} attempts to locate 3D points as natural prompts to align their projected pixel prompts across frames, while its segmentation quality heavily relies on the accuracy of 2D segmentation results. Several other works \cite{guo2023sam-graph,yin2023sai3d} follow the idea that segments each frame individually and devises a merging algorithm or graph neural network (GNN) to guide the merging process of pre-segmented superpoints. However, the limited use of 3D priors restricts the performance in both superpoints computing and region growth, which results in substandard 3D segmentation.  In this paper, we present \textbf{SA3DIP}, which incorporates more priors and constraints to minimize error accumulation and over-segmentation, by a thorough exploitation of 3D priors.

%

\section{Methodology}


\subsection{SA3DIP}
\label{SA3DIP}

\begin{figure}
	\centering
	\includegraphics[width=\linewidth]{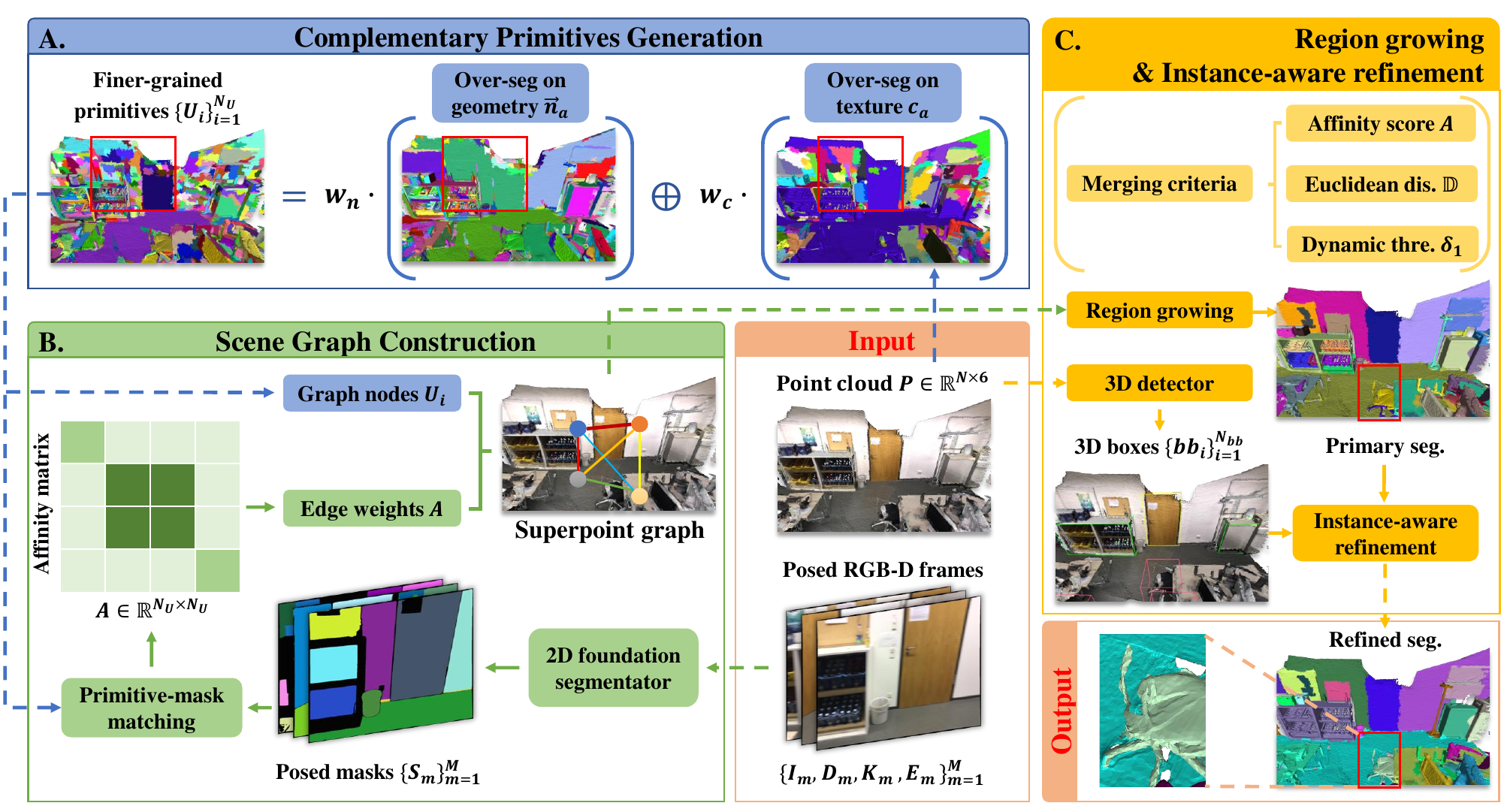}
	\caption{\textbf{Overall pipeline.} Our approach first integrates both geometric and textural priors for grouping 3D primitives (step A). Corresponding posed masks are generated using SAM. An affinity matrix is then computed based on these 2D-3D results serving as edge weights (step B). Region growing and instance-aware refinement are conducted on the constructed scene graph, utilizing 3D box constraint to address over-segmentation while maintaining the fine-grained outcomes (step C).}
	\label{figure2:pipeLine}
\end{figure}

\paragraph{Overview.}
Our overall pipeline is shown in Fig. \ref{figure2:pipeLine}. Given a point cloud $P \in \mathbb{R}^{N\times6}$ and corresponding 2D data $\{I_m,D_m,K_m,E_m\}_{m=1}^M$, which denote the RGB, depth images, camera intrinsic and extrinsic parameters, respectively, the masks of 3D instances in the scene are desired as output. First, we generate complementary 3D primitives of the given point cloud via performing 3D over-segmentation on both geometric and textural priors. Then, we construct the superpoint graph by treating the 3D primitives and their relations as nodes and edges of the graph, respectively. Leveraging the 2D masks generated by 2D foundation segmentators like SAM, we create the affinity matrix which contains the node features and edge weights. Finally, we perform affinity- and distance-aware region growing to merge the 3D primitives. Further merging is introduced by considering the supplemental prior from 3D space, which is implemented using a 3D detector.

\paragraph{Complementary primitives generation.}
Following the graph-cut algorithm in \cite{felzenszwalb2004efficient}, we calculate complementary primitives by employing over-segmentation on both geometric and textural priors. Previous methods \cite{yin2023sai3d,guo2023sam-graph} only consider the geometry information during the primitive generation process. As shown in the middle example in Fig. \ref{figure2:pipeLine}-A, under-segmentation occurs in regions where the door, wall, and board exhibit similar normals. Errors at this initial stage propagate and accumulate, adversely affecting the final segmentation. In contrast, we propose to incorporate additional textural prior, as illustrated in the right example in Fig. \ref{figure2:pipeLine}-A, which leads to finer-grained primitives. Specifically, for a 3D scene $P$, we first treat each point $p_{a} \in {P}$ as a node $v_{a}$ and calculate the edge weights $w({v}_a,{v}_b)$ for each pair of nodes $v_{a}$ and $v_{b}$. We begin by estimating the normal $\mathbf{n}_{a}$ for all $p_{a}$ using corresponding 3D coordinates $f_{a}$. Next, we extract the additional color information $c_{a}$, which previous methods fail to exploit. Note that the combination of $f_{a} \in \mathbb{R}^{3}$ and $c_{a} \in \mathbb{R}^{3}$ represents the complete point $p_{a} \in \mathbb{R}^{6}$. We compute the cosine similarity between normals $\mathbf{n}_{a}$ and $\mathbf{n}_{b}$, and normalized Euclidean distance between colors $c_{a}$ and $c_{b}$. The final edge weights $w({v}_a,{v}_b)$ are obtained by a weighted sum of these two dissimilarities:
\begin{equation}
	w({v}_a,{v}_b)=w_n \cdot \frac{\mathbf{n}_a \cdot \mathbf{n}_b}{\|\mathbf{n}_a\| \|\mathbf{n}_b\|} + w_c \cdot \sqrt{\sum_{k=1}^3 (c_{ak} - c_{bk})^2}.
\end{equation}
Subsequently, we cluster points to the finer-grained primitives $\left\{U_i\right\}_{i=1}^{N_U}$ based on each pair of $w({v}_a,{v}_b)$.

\paragraph{Scene graph construction.}
As shown in Fig. \ref{figure2:pipeLine}-B, we follow the paradigm to construct a superpoint graph for the given scene. The generated primitives serve as nodes and the affinity scores obtained through a matching algorithm serve as edge weights. Specifically, we first obtain the 2D projection $U_{i}^{m}\in \mathbb{R}^{H\times W \times2}$ of the $i$-th 3D primitive $U_{i}\in \mathbb{R}^{N_i\times3}$ on the $m$-th image by utilizing the common pinhole camera matrix:
\begin{equation}
	(U_{i}^{m},1)^{T}={K_m}\cdot{E_m}\cdot{(U_i,1)^{T}}.
\end{equation}
We feed the $m$-th RGB-image $I_m$ into the 2D foundation segmentator, e.g. SAM, to obtain its masks $S_m$. The primitive-mask matching algorithm is then performed on 2D projections $U_{i}^{m}$ and the masks $S_m$ for computing affinity scores. To be specific, we calculate a normalized histogram vector $\mathbf{e}_{i,m}$ to collect the 2D masks in $S_{m}$ covered by rendered $U_{i}^{m}$, since multiple labels may be covered due to the ambiguity or inaccuracy in 2D masks. The affinity score between $i$- and $j$-th superpoints on the $m$-th frame is obtained by computing the cosine similarity of their histogram vectors:
\begin{equation}
	A_{i,j}^m=\frac{\mathbf{e}_{i,m} \cdot \mathbf{e}_{j,m}}{\|\mathbf{e}_{i,m}\| \|\mathbf{e}_{j,m}\|}.
\end{equation}
Traversing all $M$ images yields all affinity scores between $U_{i}$ and $U_{j}$. However, primitives may not be visible in all frames, which leads to invalid affinity scores. To address this, we apply a visibility-based filter on the obtained scores. The visibility $\mathbb{V}_{i}^m\in(0,1)$ is defined as the ratio of the visible point number of $U_{i}$ on $m$-frame to the total point number of that in the scene. Thus, the final affinity score $A_{ij}$ is calculated in a weighted-sum manner as:
\begin{equation}A_{i,j}=\frac{\sum_{m=1}^M\left(\gamma_{i,j}^mA_{i,j}^m\right)}{\sum_{m=1}^M\gamma_{i,j}^m},\end{equation}
where the weight $\gamma_{i,j}^m$ is calculated as the product of $\mathbb{V}_{i}^m$ and $\mathbb{V}_{j}^m$. Iteratively processing through all superpoints and 2D frames yields the adjacency matrix $A\in\mathbb{R}^{N_U\times N_U}$. Thus, the superpoint graph of the scene is constructed with primitives $\left\{U_i\right\}_{i=1}^{N_U}$ as nodes and adjacency matrix $A$ as edge weights.

\begin{multicols}{2}
	\columnbreak
	\paragraph{Region growing and instance-aware refinement.}
	We perform affinity- and distance-aware region growing on the constructed graph. Previous methods inherit the part-level segmentation tendency from 2D masks output by SAM, often leading to over-segmentation in 3D space. For instance, the chair of the primary segmentation shown in Fig. \ref{figure2:pipeLine}-C is segmented as two distinct parts. To address this issue, we propose to exploit supplemental prior from 3D space by the integration of a 3D detector \cite{shen2023v, lu2023open, cao2024coda} for further merging. As shown in Fig. \ref{figure2:pipeLine}-Output, the constraint provided by additional 3D prior rectifies the over-segmented instances while preserving the capability to handle detailed objects.
	
	At the primary merging stage, we incorporate not only the affinity scores $A_{i,j}$ but also the Euclidean distances $\mathbb{D}_{i,j}$ between nodes $U_{i}$ and $U_{j}$, thus to 
	
	\columnbreak
	\begin{algorithm}[H]
		\caption{Instance-aware refinement}
		\label{post refine}
		\begin{algorithmic}[1]
			\State \textbf{Input:} bounding boxes $\{bb_{i}\}_{i=1}^{N_{bb}}$ in ascending order of volume, primary 3D segmentation masks $l\in \mathbb{R}^N$, threshold $\delta_{2}$
			\State \textbf{Output:} Updated 3D segmentation $l'$\\
			$O_{i}$ $\leftarrow$ $\emptyset$, $l'$ $\leftarrow$ $l$\\
			$L$ $\leftarrow$ \text{\{max instance ID in $l$}\} + 1
			\For{each $bb_{i}$ in $\{bb_{i}\}_{i=1}^{N_{bb}}$}
			\State $O_{i}$ $\leftarrow$ $l$ $\cap$ $bb_{i}$
			\For{each instance ID $O_{i}'$ in $O_{i}$}
			\State  $\sigma_{i}$ $\leftarrow$ $\frac{\text{points in } O_{i}'}{\text{points in } \{l' = O_{i}'\}}$
			\If{$\sigma_{i} > \delta_{2}$}
			\State $\{l'=O_{i}'\}$ $\leftarrow$ $L$
			\EndIf
			\EndFor
			\State $L$ $\leftarrow$ $L+1$
			\EndFor
			
		\end{algorithmic}
	\end{algorithm}
	
\end{multicols}

 introduce a certain level of global awareness. Dynamic thresholds $\delta_{1} \in \mathbb{R}^{N_{t}} $ are applied to reduce the initial erroneous merges and subsequent error accumulation. Specifically, we multiply $A_{i,j}$ with decay factor $\epsilon_\mathbb{D}$ to get the merging confident score $\delta_{i,j}$ between $U_{i}$ and $U_{j}$:
\begin{equation}
	\delta_{i,j}=\epsilon_\mathbb{D} \cdot A_{i,j}=\frac 1{\mathbb{D}_{i,j}} \cdot A_{i,j},
\end{equation}
where $\epsilon_\mathbb{D}$ is the reciprocal of the Euclidean distance. Then we compare the confident score $\delta_{i,j}$ with the threshold $\delta_1$ to judge whether to merge the nodes $U_{i}$ and $U_{j}$. Therefore, the primary segmentation results are obtained through iterating all pairs of nodes $N_{t}$ times.

We further introduce supplemental prior from 3D space by employing a detection-based instance-aware refinement. As shown in Algorithm \ref{post refine}, we gather all points $O_{i}$ within the bounding box $bb_{i}$, and assess the proportion $\sigma_{i}$ of points belong to instance ID $O_{i}'$ in $O_{i}$ to that of the entire scene $P$. If the ratio $\sigma_{i}$ exceeds a specified threshold $\delta_{2}$, it indicates high confidence that the points with instance ID $O_{i}'$ represent a portion of an over-segmented instance. We assign a new label to all points that exceed the threshold, thereby rectifying the over-segmented instance. However, there is a possibility that smaller objects which are in close proximity to or situated on larger objects are false-corrected. To mitigate this issue, we opt to pre-sort the bounding boxes based on size. In this way, the corrections are performed in descending order of bounding box size to ensure that smaller objects retain their independence in the final output.

\begin{figure}
	\centering
	\includegraphics[width=\linewidth]{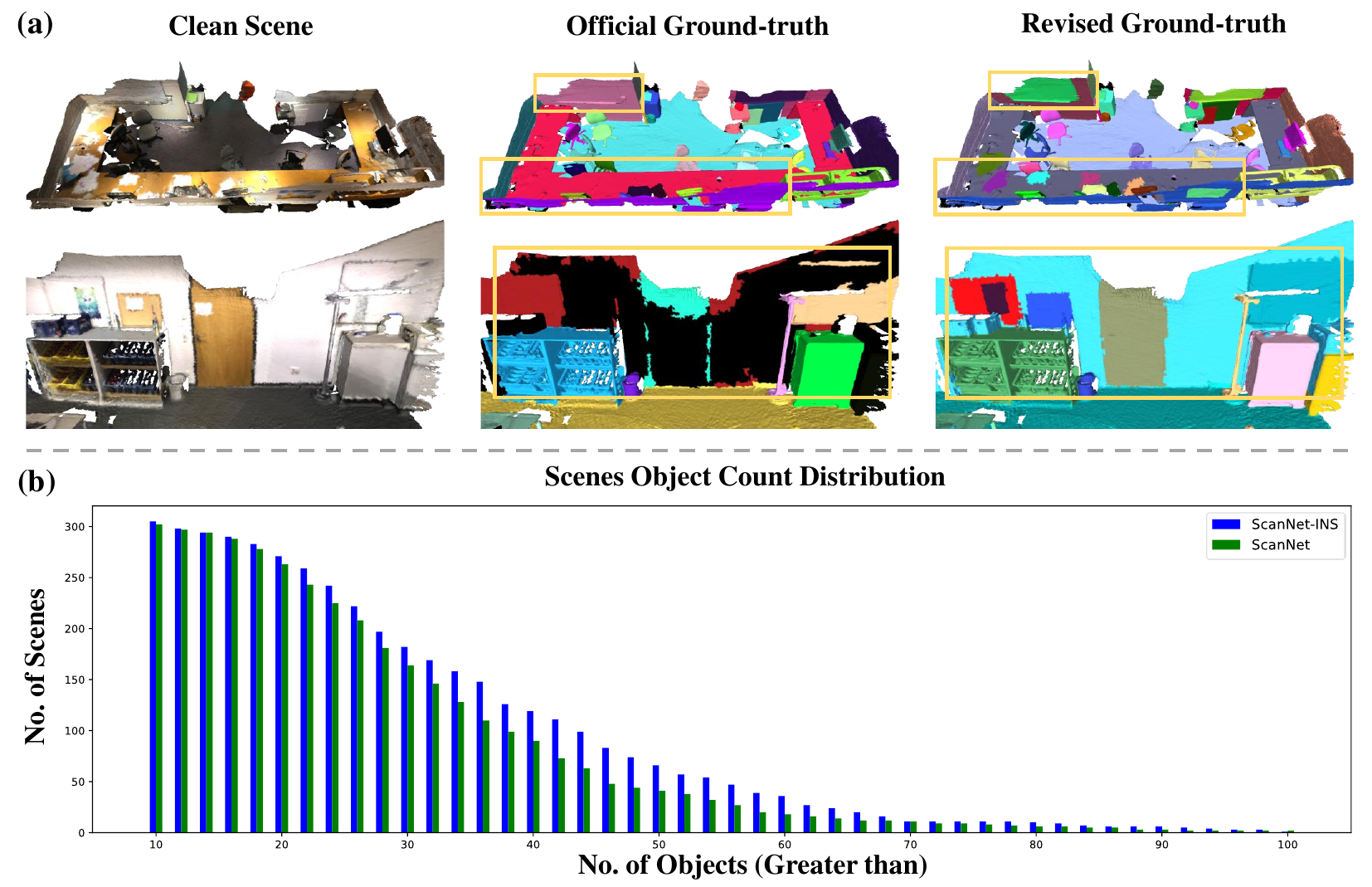}
	\caption{Overview of our proposed ScanNetV2-INS. We present the new benchmark for 3D class-agnostic instance segmentation, which rectifies incomplete annotations and incorporates more instances based on ScanNetV2. Row (a) shows the comparison before and after revision, and row (b) illustrates the object counts per scene between the two benchmarks.}
	\label{figure3:scannetv2}
\end{figure}

\subsection{ScanNetV2-INS}
\label{dataset}
ScanNetV2 has served as a standard benchmark for evaluating model performance. However, it includes a notable proportion of low-quality ground truth labels, potentially leading to misleading results. To address this issue, we introduce a refined version of the dataset, termed ScanNetV2-INS, wherein annotations are enhanced at the point level.

\paragraph{Imperfection in vanilla ScanNetV2.}
The original ScanNetV2 exhibits imperfections in its ground truth annotations, primarily manifesting in two aspects. Firstly, certain obvious instances remain unmarked. For example, as illustrated in Fig. \ref{figure3:scannetv2}-a top row, the board on the wall and papers on the desk are neglected. Secondly, some instances are incompletely annotated. For instance, doors and boards in Fig. \ref{figure3:scannetv2}-a bottom row that are clearly visible in clean point clouds feature large areas of black (indicating "unlabeled") in the annotations. The prevalence of these significantly impacts the accuracy of evaluation metrics and leads to erroneous estimations of model performance. Thus, corrective measures are imperative.

\paragraph{Revision of ScanNetV2.}

With the aid of a recently released annotation tool AGILE3D proposed in \cite{yue2023agile3d}, we perform point-level updates on the ground truth annotations for all 312 scenes in the validation set efficiently. The revision primarily addresses two aforementioned deficiencies, as shown in Fig. \ref{figure3:scannetv2}-a right column. Firstly, we re-label the instances where the ground truth was obscured by unlabeled black points, such as the door and boards. Secondly, we assign class-agnostic labels to certain instances that were clearly discernible to the human perception but were not originally annotated in the ground truth, such as the papers on the desk and the poster on the wall.

\begin{table}[t]
	\centering
	\caption{Instance number within varying point range of ScanNetV2 and ScanNetV2-INS dataset.}
	\begin{tabular}{ccccccc}
		\toprule
		\multirow{2}*{Dataset} & \multicolumn{6}{c}{Point number of the instance} \\
		\cmidrule{2-7} 
		& <500 & 500-1000 & 1000-2000 & 2000-5000 & 5000-10000 & >10000 \\
		\midrule
		ScanNetV2 & 252 & 452 & 1119 & 1690 & 567 & 284 \\
		ScanNetV2-INS & 692 & 748 & 1366 & 1873 & 626 & 291 \\
		\bottomrule
	\end{tabular}
	
	\label{tab:points}
\end{table}

\begin{multicols}{2}
	\columnbreak
	\paragraph{Statistic analysis and limitation of ScanNetV2-INS.}
	Fig. \ref{figure3:scannetv2}-b demonstrates how many scenes hold more than (10, 20, ..., 100) instances.  In Tab. \ref{tab:points} we show the number of instances with varying point counts within specified ranges for two datasets. ScanNetV2-INS dataset features more smaller objects, which requires the model to have finer-grained instance perception capabilities. As 
	
	\columnbreak
	\begin{table}[H]
		\centering
		\caption{Instance count of ScanNetV2 and ScanNetV2-INS dataset.}
		\begin{tabular}{ccccc}
			\toprule
			\multirow{2}*{Dataset} & \multicolumn{4}{c}{Instance Count} \\
			\cmidrule{2-5} 
			& Min & Max & Avg & Total \\
			\midrule
			ScanNetV2 & 2 & 47 & 14 & 4364 \\
			ScanNetV2-INS & 2 & 54 & 17 & 5596 \\
			\bottomrule
		\end{tabular}
		
		\label{tab:instance count}
	\end{table}
	
\end{multicols}

a result, the instance count of our new dataset, as shown in Tab. \ref{tab:instance count}, significantly increases. Therefore, it better reflects and poses more challenges on the model performance. However, our dataset consists of only the revised version of 312 scenes in the validation set, focusing on the evaluation use of 3D class-agnostic instance segmentation in the context of no-training methods.

\section{Experiments}
\label{others}
In this section, we quantitatively evaluate our SA3DIP on ScanNet series (including the vanilla ScanNetV2 \cite{dai2017scannet}, our ScanNetV2-INS, and more challenging ScanNet++ \cite{yeshwanthliu2023scannetpp}), Matterport3D \cite{chang2017matterport3d} and Replica \cite{straub2019replica} datasets to demonstrate its effectiveness and robustness in 3D instance segmentation. Qualitative visualizations for ScanNet series datasets are also provided for a more intuitive comparison with other methods.

\subsection{Experiment settings}
\paragraph{Datasets.}
ScanNet \cite{dai2017scannet} integrates a comprehensive array of 2D and 3D data sourced from indoor environments, facilitated by an iPad application in tandem with depth sensors. This dataset includes RGB and depth images, along with 3D point cloud data, all meticulously annotated with semantic and instance labels. It encompasses an extensive collection of over 2.5 million views derived from more than 1500 scans. In contrast, ScanNet++ \cite{yeshwanthliu2023scannetpp} represents a recently introduced indoor dataset exhibiting a similar composition to ScanNet but boasting higher-resolution 3D geometry and more detailed data annotations. ScanNet++ data is captured using advanced equipment, including the Faro Focus Premium laser scanner, iPhone 13 Pro, and a DSLR camera equipped with a fisheye lens. Our proposed ScanNet-INS encompasses a revised version of all 312 validation scenes in ScanNetV2, while maintaining consistency with ScanNetV2 in terms of data and label format. It provides a more accurate metric and fairer comparison between methods. 

\paragraph{Parameter settings.}
We conduct all experiments on a single RTX4090. The weights for geometry and texture used in the complementary primitives generation are set as $w_n=0.96$ and $w_c=0.04$. This is because texture prior such as RGB values are not robust enough when being used solely due to lighting conditions, reflections, shadows, and noise collected by sensors. We conduct detailed ablation study on the choice of the two weights in the later section. The threshold $\delta_{1}$ in the region growing is empirically set as $[0.9,0.8,0.7,0.6,0.5]$ for ScanNetV2 and ScanNetV2-INS, $[0.9,0.8,0.7]$ for ScanNet++, and the threshold $\delta_{2}$ in instance-aware refinement is set as $0.75$ experimentally.

\paragraph{Evaluation metrics.}
We evaluate the quantitative results with the widely used Average Precision score. Following \cite{guo2023sam-graph,yin2023sai3d,takmaz2024openmask3d,schult2023mask3d}, we report AP with thresholds of 25\% and 50\% (denoted as $\mathbf{AP}_{25}$, $\mathbf{AP}_{50}$) and averaged over all overlaps between [50\% and 95\%] at 5\% steps ($\mathbf{mAP}$). Since the 2D foundation segmentation model produces class-agnostic masks, we ignore semantic class labels in the evaluation and consider only the accuracy of the instance masks themselves.

\paragraph{Baselines.}
We compare our approach with both closed-vocabulary and open-vocabulary methods. Mask3D \cite{schult2023mask3d} trained on ScanNetV2 serves as the closed-vocabulary baseline. Recent methods based on leveraging 2D foundation models, including SAM3D \cite{yang2023sam3d} (with and without ensemble process), SAM-graph \cite{guo2023sam-graph}, SAI3D \cite{yin2023sai3d}, and SAMPro3D \cite{xu2023sampro3d} are compared as open-vocabulary methods. In addition, we compare with the traditional point grouping method proposed by Felzenszwalb \cite{felzenszwalb2004efficient}.


\subsection{Results on ScanNet series}

Tab. \ref{comparisonresult} shows the quantitative results of our approach in comparison with other methods on ScanNetV2, ScanNetV2-INS, and ScanNet++ datasets. Our method achieves the best performance among all three datasets, showing the effectiveness of our approach. Specifically, our SA3DIP outperform 7.9\% $\mathbf{mAP}$, 8.4\% $\mathbf{AP}_{50}$, and 6.0\% $\mathbf{AP}_{25}$ on ScanNetV2, 3.6\% $\mathbf{mAP}$, 3.8\% $\mathbf{AP}_{50}$, and 2.9\% $\mathbf{AP}_{25}$ on ScanNetV2-INS. On the challenging ScanNet++, our method still obtains 2.5\% $\mathbf{mAP}$, 2.7\% $\mathbf{AP}_{50}$, and 2.0\% $\mathbf{AP}_{25}$ gain. Note that all methods except SAM3D experience a drop in precision on ScanNetV2-INS dataset compared with the vanilla ScanNetV2. This indicates that our proposed ScanNetV2-INS poses more challenges in identifying fine-grained objects and yields fairer metrics. SAM3D, due to its limited segmentation capability, tends to produce a significantly higher number of instances than the actual objects in the scene. Consequently, its metrics do not show noticeable changes on finer-grained ScanNetV2-INS.

%

\begin{figure}
	\centering
	\includegraphics[width=\linewidth]{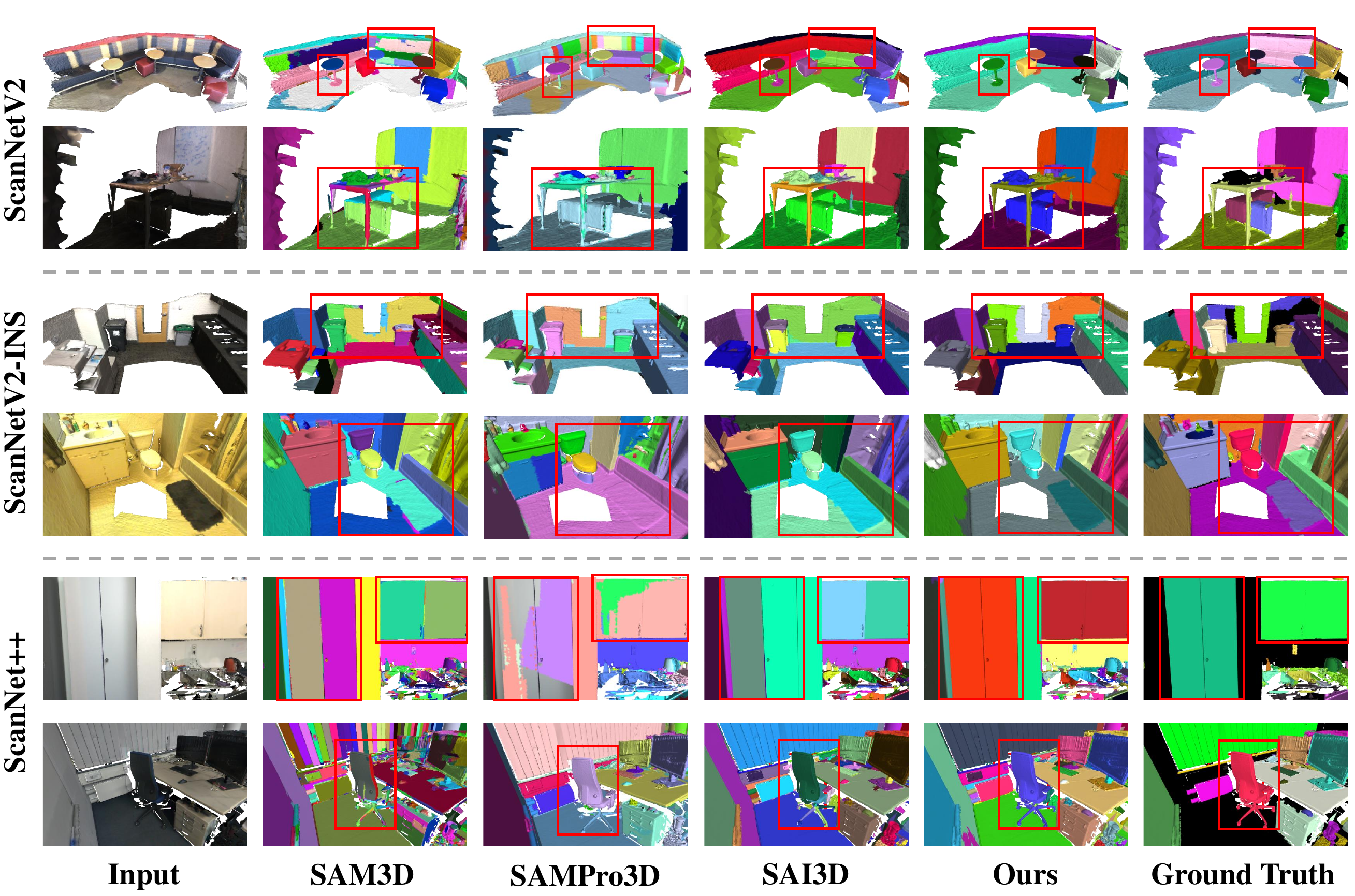}
	\caption{Visual comparison between our method with SAM3D \cite{yang2023sam3d}, SAMPro3D \cite{xu2023sampro3d}, and SAI3D \cite{yin2023sai3d} on ScanNetV2, ScanNetV2-INS, and ScanNet++ dataset. Among all datasets, our method shows the most robust and accurate segmentation.}
	\label{figure4:result}
\end{figure}

We also present qualitative results in Fig. \ref{figure4:result}. The visual comparison further proves the effectiveness of our method. As shown in the first two rows of Fig. \ref{figure4:result}, our method maintains a better instance awareness and is capable of identifying the tables as a whole. Moreover, by utilizing more accurate 3D primitives, our approach is the only one able to segment the door out from the wall, as shown in the third row of Fig. \ref{figure4:result}. This showcases the significance of exploiting the potential 3D priors.

\begin{table}[t]
	\centering
	\caption{Class-agnostic 3D instance segmentation comparison on ScanNetV2, ScanNetV2-INS, and ScanNet++ datasets.}
	\begin{tabular}{cccccccccc}
		\toprule
		\multirow{2}*{Method} & \multicolumn{3}{c}{ScanNetV2} & \multicolumn{3}{c}{ScanNetV2-INS} & \multicolumn{3}{c}{ScanNet++} \\ \cmidrule{2-10} 
		& $\textbf{mAP}$ & $\textbf{AP}_{50}$ & $\textbf{AP}_{25}$ & $\textbf{mAP}$ & $\textbf{AP}_{50}$ & $\textbf{AP}_{25}$ & $\textbf{mAP}$ & $\textbf{AP}_{50}$ & $\textbf{AP}_{25}$ \\
		\midrule
		\textit{Closed-vocabulary} \\
		Mask3D \cite{schult2023mask3d} & 31.1 & 44.9 & 58.0 & 29.1 & 43.9 & 56.3 & 9.9 & 17.3 & 25.8 \\
		\cmidrule{1-10} 
		\textit{Open-vocabulary} \\
		Felzenszwalb \cite{felzenszwalb2004efficient} & 5.0 & 12.7 & 38.9 & 2.8 & 6.5 & 24.0 & 4.1 & 9.2 & 25.3 \\
		\makecell{SAM3D \cite{yang2023sam3d} \\ (w/o ensemble)} & 12.4 & 28.7 & 57.4 & 12.5 & 28.9 & 57.8 & 1.1 & 4.5 & 15.4 \\
		\makecell{SAM3D \cite{yang2023sam3d} \\ (w/ ensemble)} & 20.1 & 33.3 & 52.1 & 20.0 & 33.2 & 52.2 & 7.2 & 14.2 & 29.4 \\
		SAM-graph \cite{guo2023sam-graph} & 24.1 & 40.3 & 65.9 & 23.1 & 39.5 & 64.1 & 12.9 & 25.3 & 43.6 \\
		SAI3D \cite{yin2023sai3d}& 30.8 & 50.5 & 70.6 & 28.9 & 49.2 & 69.7 & 17.1 & 31.1 & 49.5 \\
		SAMPro3D \cite{xu2023sampro3d} & 33.7 & 56.2 & 75.3 & 32.5 & 54.8 & 73.4 & 18.9 & 33.7 & 51.6 \\
		\cmidrule{1-10} SA3DIP (ours) & \textbf{41.6} & \textbf{64.6} & \textbf{81.3} & \textbf{36.1} & \textbf{58.6} & \textbf{76.3} & \textbf{21.4} & \textbf{36.4} & \textbf{53.6} \\
		\bottomrule
	\end{tabular}
	
	\label{comparisonresult}
\end{table}

\begin{table}[t]
	\centering
	\caption{Ablation studies on prior we exploit with varying weights.}
	\begin{tabular}{ccccccccc}
		\toprule
		\multirow{2}*{$w_n$} & \multirow{2}*{$w_c$} & \multirow{2}*{3D Space Prior} & \multicolumn{3}{c}{ScanNetV2} & \multicolumn{3}{c}{ScanNetV2-INS} \\ \cmidrule{4-9} 
		&&& $\textbf{mAP}$ & $\textbf{AP}_{50}$ & $\textbf{AP}_{25}$ & $\textbf{mAP}$ & $\textbf{AP}_{50}$ & $\textbf{AP}_{25}$ \\
		\midrule
		1 & 0 & × & 30.8 & 50.5 & 70.6 & 28.9 & 49.2 & 69.7 \\
		0 & 1 & × & 10.4 & 18.1 & 32.5 & 9.5 & 17.0 & 31.1 \\
		0.4 & 0.6 & × & 27.3 & 47.4 & 69.8 & 25.6 & 46.3 & 69.4  \\
		0.96 & 0.04 & × & 29.3 & 49.2 & 70.5 & 27.4 & 48.3 & 70.4 \\
		\cmidrule{1-9}
		1 & 0 & \checkmark & 40.8 & 63.6 & 80.7 & 35.9 & 57.8 & 75.4 \\
		0 & 1 & \checkmark & 12.7 & 22.1 & 37.2 & 11.0 & 19.7 & 34.1 \\
		0.4 & 0.6 & \checkmark & 39.1 & 62.7 & 80.2 & 33.5 & 56.3 & 75.0 \\
		0.96 & 0.04 & \checkmark & \textbf{41.6} & \textbf{64.6} & \textbf{81.3} & \textbf{36.1} & \textbf{58.6} & \textbf{76.3} \\
		\bottomrule
	\end{tabular}
	
	\label{tab:ablation}
\end{table}

\subsection{Ablation studies}
We conduct detailed ablation studies on prior with varying weights $w_n$ and $w_c$. We report the metrics in Tab. \ref{tab:ablation}. We assigned several weights for geometry and texture to test their contribution. Specifically, we conduct one with configuration of $w_n=0.4$ and $w_c=0.6$ which yields the similar number of 3D primitives as the primitives used in SAM3D, SAI3D and others, for a fair comparison. 

It can be observed that texture prior are not robust enough when being used solely due to the influence of shadows, reflection and so on. Therefore, we choose to assign less weight to the textural prior, thus to exploit it while minimizing its negative impact. The experiments show that the setting with $w_n=0.96$ and $w_c=0.04$ suits best for our approach. 

However, it is noticed that incorporating only complementary primitives yields a slight drop in average precision on both datasets. It is related to the definition of the metric AP (ratio of correctly identified instances to the total number of identified instance). This AP metric is more in favor of under-segmentation rather than over-segmentation, since the former yields high precision and fewer false positive cases, while the latter gives fewer precision and higher recall.

\subsection{More experiments}
We conduct further experiments and corresponding ablation studies on both Matterport3D \cite{chang2017matterport3d} and Replica \cite{straub2019replica} dataset to test the robustness and generalization ability of our method. Matterport3D dataset contains 194,400 RGB-D images of 90 building-scale indoor scenes and exhibits more view changes on its 2D frames compared to ScanNet. Replica dataset incorporates 18 highly photo-realistic 3D indoor scene reconstructions with dense geometry, high resolution and dynamic range textures. Our method clearly gives better quantitative results, as shown in Tab. \ref{tab:replica}.

\begin{table}[t]
	\centering
	\caption{Experiments on Matterport3D and Replica datasets.}
	\begin{tabular}{ccccccc}
		\toprule
		 & $w_n$ & $w_c$ & 3D Space Prior & $\textbf{AP}$ & $\textbf{AP}_{25}$ & $\textbf{AP}_{50}$ \\
		\hline
		\textbf{Matterport3D dataset} \cite{chang2017matterport3d} \\
		OpenMask3D \cite{takmaz2024openmask3d} & / & / & × & 15.3 & 28.3 & 43.3 \\
		OVIR-3D \cite{lu2023ovir} & / & / & × & 6.6 & 15.6 & 28.3 \\
		SAM3D w/ ensemble \cite{yang2023sam3d} & 1 & 0 & × & 10.1 & 19.4 & 36.1 \\
		SAI3D \cite{yin2023sai3d} & 1 & 0 & × & 18.9 & 35.6 & 56.5 \\
		\midrule
		\textbf{Ablation of ours} & & & & & & \\
		Ours \#1 & 1 & 0 & \checkmark & 19.8 & 36.6 & 56.2 \\
		Ours \#2 & 0.9 & 0.1 & × & 18.1 & 35.7 & \textbf{62.3} \\
		Ours \#3 & 0.9 & 0.1 & \checkmark & \textbf{20.6} & \textbf{38.3} & \textit{61.0} \\
		\bottomrule
		
		\textbf{Replica dataset} \cite{straub2019replica}  \\
		SAM3D w/o ensemble \cite{yang2023sam3d} & / & / & × & 11.9 & 22.9 & 38.4 \\
		SAM3D w/ ensemble \cite{yang2023sam3d} & 1 & / & × & 12.4 & 20.0 & 32.0 \\
		SAMPro3D \cite{xu2023sampro3d} & 1 & 0 & × & 13.1 & 25.2 & 44.7 \\
		SAI3D \cite{yin2023sai3d} & 1 & 0 & × & 20.4 & 30.7 & 42.9 \\
		\midrule
		\textbf{Ablation of ours} & & & & & & \\
		Ours \#1 & 1 & 0 & \checkmark & 21.0 & 31.5 & 43.6 \\
		Ours \#2 & 0.9 & 0.1 & × & 20.8 & 32.8 & 46.2 \\
		Ours \#3 & 0.9 & 0.1 & \checkmark & \textbf{22.6} & \textbf{34.2} & \textbf{47.1} \\
		\bottomrule
	\end{tabular}
	
	\label{tab:replica}
\end{table}

\subsection{Limitations}
Due to the trade-off between efficiency and accuracy, we choose to compute 3D superpoints based on only 3D priors. This yields an extremely short time of execution within a few seconds, while it may lead to an overwhelming number of superpoints which introduces challenges in the merging process. Moreover, for high-resolution point clouds with vivid light and shade effects, the superpoints generated based on geometric and texture is not enough yet. One approach is to design a more sophisticated pre-segmentation model with semantic awareness. Besides, though constraints provided by 3D prior are introduced, the affinity matrix based on 2D masking still relies heavily on the accuracy of 2D foundation segmentators. Designing a more robust merging algorithm or better leveraging various 2D foundation models shows promise in the future.

\section{Conclusion}
In this paper, we introduce a novel method for segmenting any 3D instances by exploiting the potential 3D priors. The key idea is to incorporate more 3D priors into the 2D foundation model guided pipeline and leverage not only knowledge transferred from 2D space but also features in 3D space. We first generate complementary 3D superpoint primitives based on both geometric and textural priors to reduce the initial errors that accumulate in subsequent procedures. Then we introduce supplemental constraints from the 3D space by using a 3D detector. Along with the constructed affinity matrix by using 2D masks, the region growing and refinement process is performed on the 3D primitives. Furthermore, we propose ScanNetV2-INS with complete ground truth labels and supplement additional instances for 3D class-agnostic instance segmentation, which produces unbiased metrics on comparing different methods. Experimental evaluations on ScanNetV2, ScanNetV2-INS, and ScanNet++ datasets demonstrate the effectiveness of our approach. We believe that we pioneer at exploiting the importance of 3D priors in the 2D foundation model guided pipeline, and it should draw attention toward future research that methods trying to extend 2D foundation models into 3D space should not overlook the role of inherent 3D priors.

\section*{Acknowledgments}
This work was supported in part by the National Natural Science Foundation of China under Grant 62372348, Grant 62441601, Grant 62176195, Grant U21A20514; in part by the Key Research and Development Program of Shaanxi under Grant 2024GX-ZDCYL-02-10; in part by Shaanxi Outstanding Youth Science Fund Project under Grant 2023-JC-JQ-53; in part by the Fundamental Research Funds for the Central Universities under Grant QTZX24080.
%
%

{
\small

\bibliography{references}

}


%
%
%
%
%
%
%
%
%
%


\end{document}